\documentclass{article}
\usepackage[utf8]{inputenc}
\usepackage{multicol}
\usepackage{geometry}
\usepackage{lipsum} 
\usepackage{float} 
\usepackage{svg}
\usepackage{amsmath}
\usepackage{hyperref}
\usepackage{booktabs}
\usepackage{amsfonts}
\usepackage[square,sort,comma,numbers]{natbib}
\usepackage{algorithm2e}

\title{IPP}
\author{Zhenhao Gu}
\date{\today}

\usepackage{natbib}
\usepackage{graphicx}

\begin{document}

\section*{Abstract}
With the continuous development of machine learning technology, major e-commerce platforms have launched recommendation systems based on it to serve a large number of customers with different needs more efficiently. Compared with traditional supervised learning, reinforcement learning can better capture the user's state transition in the decision-making process, and consider a series of user actions, not just the static characteristics of the user at a certain moment. In theory, it will have a long-term perspective, producing more effective recommendation. The special requirements of reinforcement learning for data make it need to rely on an offline virtual system for training. Our project mainly establishes a virtual user environment for offline training. At the same time, we tried to improve a reinforcement learning algorithm based on bi-clustering to expand the action space and recommended path space of the recommendation agent.

\section*{Keywords}
Recommendation Systems, Deep Reinforcement Learning, Bi-clustering Algorithm

\begin{titlepage}
   \begin{center}
       \vspace*{1cm}

       \Huge{\bf Interactive Search Based on\\ Deep Reinforcement Learning}

       \vspace{0.5cm}
       \Large{19th IPP Project}
            
       \vfill

       \textbf{Instructor}: Prof. Paul Weng, Joint Institute\\
       \textbf{Author}: Yu Yang, Gu Zhenhao, Tao Rong, Ge Jingtian, Chang Kenglun

       \vfill

       \includegraphics[width=0.3\textwidth]{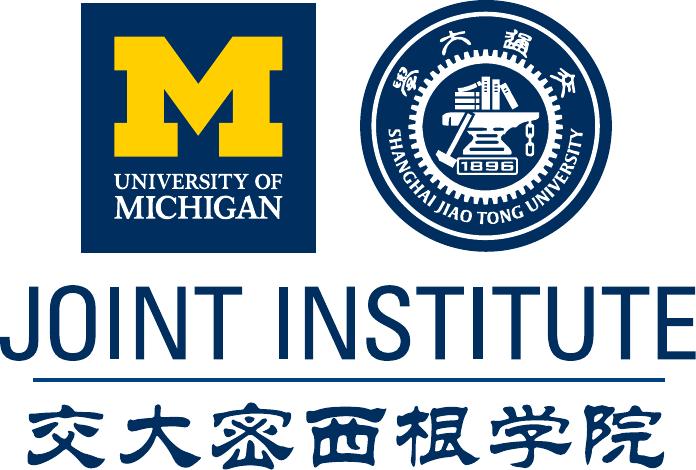}
        \vfill
       \today
            
   \end{center}
\end{titlepage}

\newpage
\tableofcontents
\newpage

\section{Introduction}
With the rise of online shopping, hotel recommendation platform, movie recommendation platform, etc., recommendation system is becoming more and more important and profitable. Using recommendation system, we can put the most suitable item for a particular user at the front, thus motivating the user to click on the item. Effective and accurate recommendation system is critical to boost sales and profit.

Typical recommendation system algorithms involve collaborative filtering, which can filter out the best items for a user based on the view history of all the users. This algorithm is effective, but it suffers from two problems:
\begin{itemize}
    \item The algorithm is based on a user-item matrix, which can be very large if the number of user and the number of item is large.
    \item For unobserved users (new users with no previous history), we can hardly predict the preference.
\end{itemize}
To solve these problem, we want to use deep reinforcement learning, where the agent is able to capture the users' preference from action sequence and state transition, thus acquiring better reward in the long run.


\section{Related Work}
Within the field of recommendation systems, the most common, commonly used methods that give us the best results are mostly based on deep learning methods. There are various variations and improvements on the implementation of algorithms and neural networks in the field. Most of them follow the same approach - they encode and parse information into different forms and put it into the output of a neural network. For example, in the 2019 Trivago recsys challenge, all of the top 5 teams used deep learning to assist with different ways of understanding and interpreting data and building networks, and most of them achieved an average inverse ranking of around 0.67. However, this purely deep learning approach to neural network construction is not the only way to achieve the best results.

This purely deep learning neural network does not fully exploit the sequential nature of user operations: users perform a sequence of operations to eventually make a choice, and the deep learning tend to compress an entire sequence of user operations into a single input no matter how long the user's record of operations is. This type of coding makes it difficult to efficiently compile features from a user with 1 operation and a user with 200 operations. The advantage of deep reinforcement learning is that we can take advantage of the sequential characteristics of the user's operations so that they can be interpreted more efficiently.

\subsection{Virtual Taobao}
Virtual Taobao \cite{virtualtaobao} is a method proposed to combine Reinforcement Learning with online shopping platform, Taobao. The idea of Virtual Taobao is to generate virtual customers and interactions based on real data from Taobao.

Virtual Taobao uses the following approaches

\begin{itemize}
    \item \textsc{GAN-for-Simulating-Distribution (GAN-SD)} approach to simulate customers, which adopts an extra distribution constraint to generate diverse customers.
    \item \textsc{Multi-agent Adversarial Imitation Learning (MAIL)} approach to simulate interactions, which learns customers' policies and the platform policy simultaneously.
    \item \textsc{Action Norm Constraint (ANC)} strategy to reduce the over-fitting caused by a powerful algorithm in Reinforcement Learning.
\end{itemize}

Virtual Taobao has many innovations in building user models and simulating user interactions, and is better able to simulate the Markov chain of user search $\rightarrow$ site recommendations $\rightarrow$ user reactions to help reinforcement learning agents. However, there are still many problems with virtual Taobao.
\begin{enumerate}
    \item Virtual Taobao does not specify the items. In virtual Taobao, the reinforcement system agent needs to accept an 88-dimensional user feature vector and output a 27-dimensional vector with one dimension between $-1$ and $1$ as the recommended item vector.
    \item Virtual Taobao uses static learning. The actions generated by reinforcement learning have no effect on the user's next round of actions.
\end{enumerate}

\subsection{Recsim}
Recsim \cite{recsim} is a configurable simulation platform for recommender systems make by Google, which utilized the document and user database directly. We can break Recsim into two parts,
\begin{itemize}
    \item The environment consists of a user model, a document (item) model and a user-choice model. The user model samples users from a prior distribution of observable and latent user features; the document model samples items from a prior over observable and latent document features; and the user-choice model determines the user's response, which is dependent on observable document features, observable and latent user features.
    \item The \textsc{SlateQ} Simulation Environment, which uses the \textsc{SlateQ} Algorithm to return a slate of items back to the simulation environment.
\end{itemize}
Unlike virtual Taobao, Recsim has a concrete representation of items, and the actions returned by the reinforcement learning agent can be directly associated with items. However, the user model and item model of Recsim are too simple, and without sufficient data support, the prior probability distribution for generating simulated users and virtual items is difficult to be accurate.

\section{Problem Setting}
\subsection{The Trivago Dataset}
The RecSys Challenge 2019 will be organized by trivago, TU Wien, Polytechnic University of Bari, and Karlsruhe Institute of Technology, and presents a real-world task in the travel metasearch domain. Users that are planning a business or leisure trip can use Trivago’s website to compare accommodations and prices from various booking sites. Trivago provides aggregated information about the characteristics of each accommodation to help travelers to make an informed decision and find their ideal place to stay.
\subsubsection{Data Overview}
\begin{figure}[H]
\centering
\includegraphics[width=0.8\textwidth]{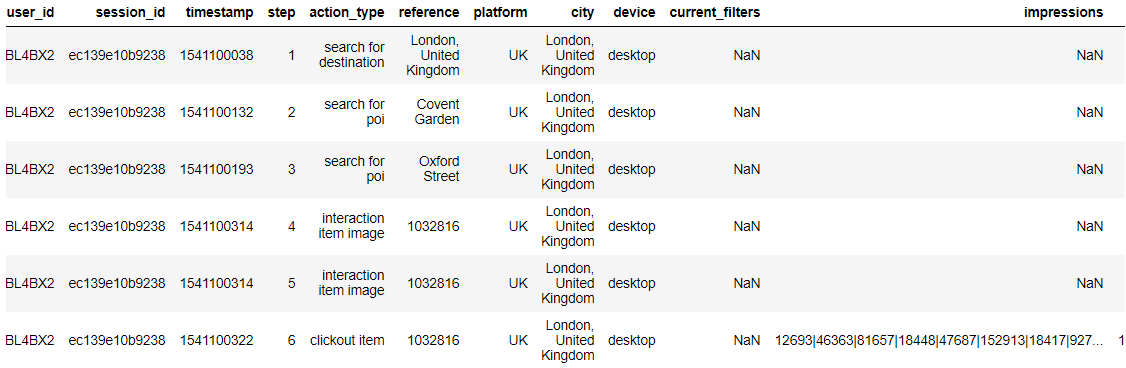}
\caption{Sample Data of Trivago Data Set.}
\label{fig:sampleData}
\end{figure}
    
    This is a typical trivago user, he will first browse some information and set up some filters and place of interest, and finally he will make his decision (The bottom line). He will be choosing from the list of hotels in the right column and the hotel he have chosen is the "reference" in that step 
    
    Our task is to predict the reference in the clickout-item step. In the test set, the reference will be hidden, and we will need to sort the list of hotels given in descending likelihood of being clicked. The final measure of our algorithm is the mean reciprocal rank (MRR). So if hotel "1032816" is the $5^{th}$ in our list. We will recieve 0.2 points in this user section. The MRR is simply the mean of all user sections.
 \begin{figure}[H]
\centering
\includegraphics[width=0.8\textwidth]{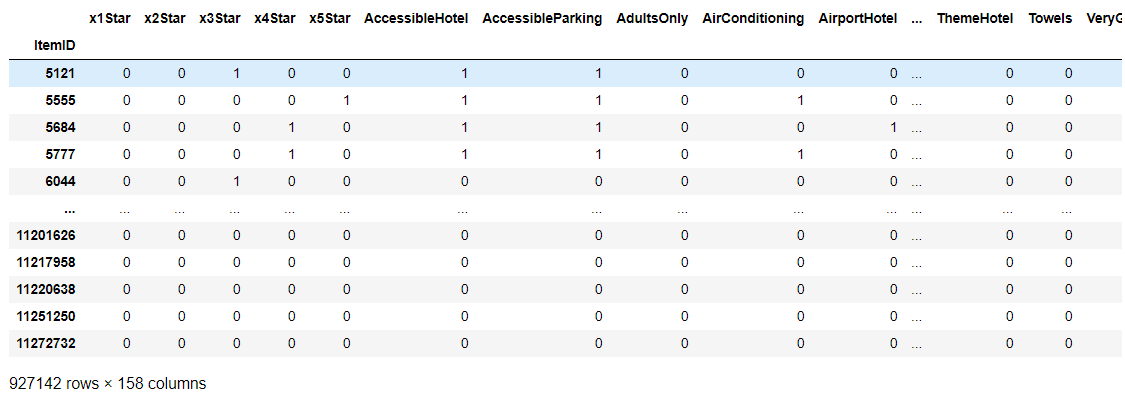}
\caption{Hotel Information.}
\label{fig:sampleData}
\end{figure}
    And here are the hotel information that is provided. We received a 156 boolean vector for each hotel, and we added 2 that is the price and the number of clicks it received. 
    
\section{Recommendation System with Deep RL}
\subsection{Data Preprocessing}
We discretize the user operation in a very special way. The state vector is encoded in a very special way. The first part of the state is a $20 \times 25$ array. Where 20 is the memory length, and 25 is the maximum number of hotels. We first flip the user operation upside down, such that the last user operation will be processed the first. Then, we'll index the the hotel in the list which we will choose the most likely hotel from from 1 to 25. Only 1 slot will be 1 in the one-hot $20 \times 25$ array, and that is the hotel which we've view in the last user operation(now it is in the first row). We will keep this process until the memory length is achieved or we ran out of user operations.

For example of the user session in \ref{fig:sampleData}. The reference in step 6 will be hidden, so we will only have the impression, namely the list of hotels and the 5 steps above. Let's say hotel "1032816" is the $10^{th}$ hotel in the list. So the first 25 slots in the $20 \times 25$ one-hot array will be 000...1...0, where the only 1 is $10^{th}$ slot. It means that, in the last operation, the user checked the $10^{th}$ hotel in the list. And for the forth step, since it has the same reference as the fifth, we'll not renew our next 25 slots and wait until a different hotel other than the hotel that is just recorded. In this case, there are none.

Since user operation can have hundreds of steps and dozens of operation on the same hotel, so we cluster the consecutive hotels and view them as 1 operation. What's more, we only keep record of the last 20 cluster operations(which covers the almost all the user sessions). And we keep the last 20 operations instead of the first 20 operations because later operations are a lot more important than the operations steps ago.

In this way, we are able to preserve much of the information provided by the user operation, even the index of the hotel list provided (The first in the hotel list is more preferable). 

Then, we'll encode the hotel state in a similar way. We will glue a $156 \times 25$ array after the user operation, which is also one-hotted. The first 156 slots will correspond to the boo leans in Figure 2, and so on and so forth.

Then, we finished building our state which is adopted from both the user operation and hotel features.

\section {Reinforcement learning environment}
In order to make a connection between our dataset and the reinforcement learning agent, we need to create a training environment. Virtual Taobao, as well as Recsim, introduced earlier, is based on establishing a distribution of user responses to randomly generate the next user action, namely
$$F_{a_n\mid a_{n-1},a_{n-2},\dots,a_0}$$
where $F$ is the user feedback action distribution and $a_i$ represents the user's action at the $i$th moment.
We argue that in the case of small amounts of data, the mathematical model of the probability distribution built may not be accurate enough because of too many variables; in the case of sufficiently large amounts of data, direct data extraction is close to the results of randomly generating data using the probability distribution. Therefore, we attempted to randomly extract data directly from the database and train a single user interaction as a step in the training of the reinforcement learning algorithm.

\begin{figure}[htbp]
    \centering
    \includegraphics[width=0.9\textwidth]{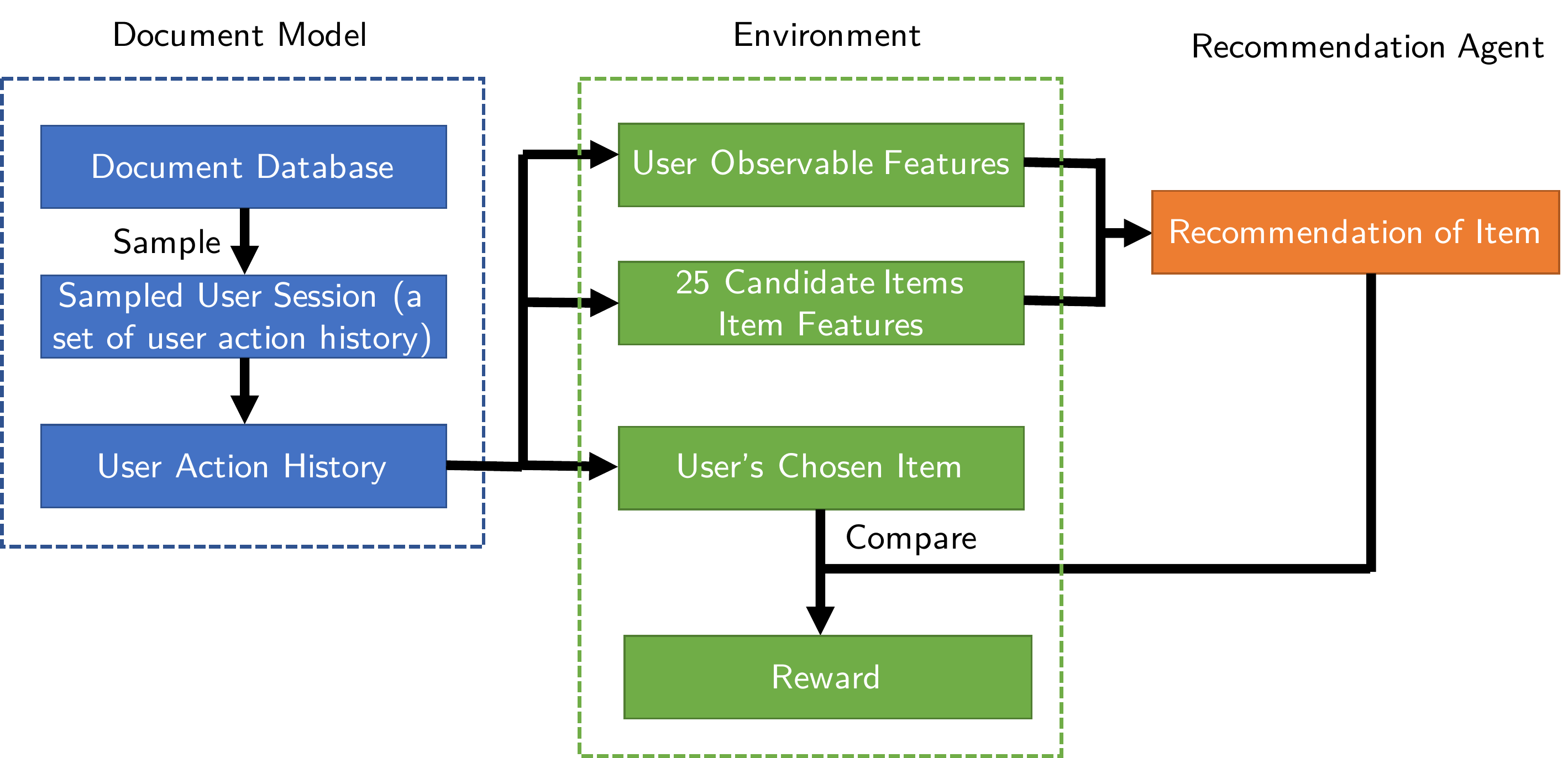}
    \caption {enhanced learning environment}
    \label{fig:env}
\end{figure}

Our reinforcement learning environment can be represented by the graph \ref{fig:env}.
This environment can be divided into three parts. \textbf{Database and Item Model} is used to continuously extract the user interaction history from the database and input it into the reinforcement learning training environment. \textbf{Environment} is used to extract basic user information and preferences from the data and interact with the \textbf{Recommendation Agent} to get the item recommended by the agent. The environment compares this item with the items actually clicked by the user in the database and generates a reward to be returned to the RL agent, which forms the training of the reinforcement learning algorithm.
\subsection {Overview}
\begin{figure}[htbp]
    \centering
    \includegraphics[width=0.9\textwidth]{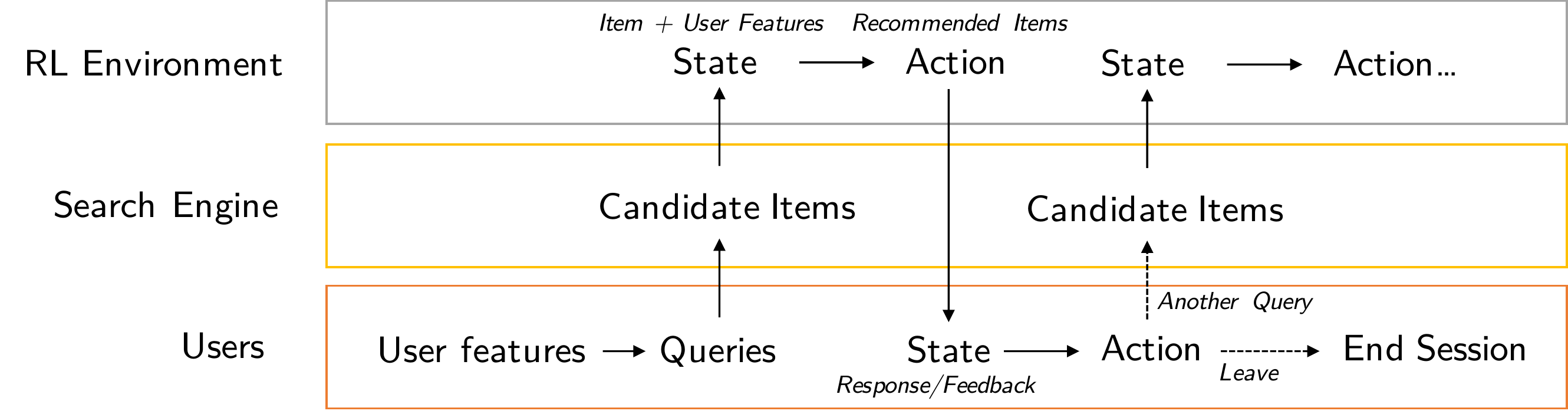}
    \caption {the Markov chain created by the reinforcement learning environment}
    \label{fig:mc}
\end{figure}

With this reinforcement learning environment, we are able to create a Markov chain, as shown in \ref{fig:mc}.
where the user's Query (i.e., one search), State (containing a response to a recommendation from the recommendation system), and Action (continue to the next search or leave) are randomly obtained directly from the dataset. This process is equivalent to randomly generating a user from a distribution of users. In the following, we will step-by-step dissect the elements of this Markov chain in the RL environment.
\subsection{State of the environment}
The state of the environment $\mathcal{S}=[u;\hat{u};i_1;i_2;\dots;i_{25}]$ is a list of user characteristics $[u;\hat{u}]$, 25 candidate items, and their corresponding characteristics $i_1,i_2,\dots,i_{25}]$.
\begin{enumerate}
    \item Basic information $u$ visible to the user, such as the platform the user is using (mobile, PC, etc.), the current filtering criteria set by the user, etc. This information is displayed as a boolean vector. This information is represented as a Boolean vector.
    
    \item The user's preference vector $\hat{u}\in \mathbb{R}^m$ has the same length as the item characteristics vector $i$. Here we derive empirical estimates of user preferences based on the user's past browsing history and an exponential moving average method. At point in time $t$, for all items $i_1,i_2,\dots,i_t$ that the user has interacted with in the past, there are
    $$\hat{u}_t=(1-\alpha)\hat{u}_{t-1}+\alpha i_{t-1}$$
    where $\alpha$ is the learning rate we define. Using this method, we are able to estimate user preferences more accurately based on the fact that they are only relevant to their most recent interactions.
\end{enumerate}
\subsection {Action of RL agent}
In our system, a recommendation is generated by the action of a reinforcement learning agent. It can take one of the following three forms.
\begin{enumerate}
    \item A user hidden preference vector $u'\in\mathbb{R}^m$ that has the same length as the item feature vector $i$. Unlike the empirical estimation method of $\hat{u}$, $u'$ is a user preference learned from the reinforcement system. We are able to calculate the user's preference for each item as $u'^T i$ from this vector.
    \item A recommended item that is one of the 25 candidate items.
    \item A vector containing 25 candidate items and ranked by the probability of being clicked by the user.
\end{enumerate}
For different forms of actions, we used different reinforcement learning algorithms.
\subsection {Reward }
Based on the items $i_{real}$ actually chosen by the user and the items recommended by the reinforcement learning algorithm $i_{recommend}$, our algorithm is able to calculate the user's feedback as a reward and train the algorithm.
\begin{enumerate}
    \item If the RL agent recommends only one item, we use the Click Through Rate :
    $$R=\frac{1}{n}\sum_{k=1}^n I(i_{real,k}=i_{recommend,k})$$
    as a reward. where the value of $I(i_{real,k}=i_{recommend,k})$ is 1 when $i_{real,k}$ is equal to $i_{recommend,k}$ and 0 otherwise. at the beginning of training, the value of $r$ may be 0 in most cases, making training slower.
    \item If the RL agent returns an arrangement of user preference vectors or candidate items, we can calculate the ranking of $r$ of the items actually chosen by the user in this arrangement. We can use the Mean Reciprocal Rank (MRR), the
    $$R= \frac{1}{n}\sum_{k=1}^n \frac{1}{r_k}$$
    as a reward. The advantage of using this reward is that RL agent will always receive $R>0$ as a reward, which can speed up training.
\end{enumerate}

\section{Deep RL Agents}
\subsection{Agent for trivago}
In our project, all agents will be fed with the same state (4.1) as input and is tested with the same random seed for several times. 
\subsubsection{DQN}
Deep-Q network is a commonly used method in deep reinforcement learning. Instead of calculating the value of each state in the case of value iteration, Q-learning methods calculates the value of each action. The benefit of Q-learning is that, you no longer need the transition function which values-based method requires. The difference between Q-learning and deep Q is that, deep Q uses a neural net work to calculate the Q-value of each action. The best property of Q learning is that it guarantees optimality. However, it is unstable and can fail miserably when the policy function is too complex to learn.

\subsubsection{Trust Region Policy Optimization}
Trust region policy optimization is a variant of the policy gradient descent algorithm \cite{trpo}. Ordinary strategy gradient descent moves to the deepest part of the gradient. However, this approach still suffers from the inability to determine step size. A compensation that is too large can have disastrous consequences for training, while a step size that is too small can make learning less efficient. Trust Region Policy Optimization exploits the improvement of the trust domain to regulate the strategy of the
\begin{eqnarray*}
\text{maximize}_{\theta}~ \mathcal{L}(\theta,\theta_{\text{old}})=\text{maximize}_{\theta}~E_t\left[\frac{\pi_{\theta}(a_t\mid s_t)}{\pi_{\theta_{\text{old}}}}A_t\right]\\
\text{subject to }E_t[\text{KL}[\pi_{\theta_{\text{old}}},\pi_{\theta}]]\leq \delta
\end{eqnarray*}

Where $\mathcal{L}(\theta,\theta_{\text{old}})$ uses data from the old policy $\pi_{\text{old}}$, the new policy $\pi_{\text{old}}$ performs relative to the old policy, and $\text{KL}[\pi_{\theta _{\text{old}},\pi_{\theta}]}$ represents the KL distance of the new policy $\pi_{\theta_{\text{old}}}$ from the old policy when using data from the old policy $\pi_{\theta_{\text{old}}}$.

TRPO is a stable method and ensures monotonically incremental progress.

\section{Results Obtained by Our RL Environment}
After completing the creation of the reinforcement learning environment, we tried several different combinations of state, action, and reward, trained them with different reinforcement learning algorithms, and compared their performance. We used both Trust Region Policy Optimization and deep Q network algorithms.

We compared the different algorithms using different RL algorithms, as shown in table \ref{tab:comp}.
\begin{table}[htbp]
    \centering
    \begin{tabular}{lll}
    \toprule
         action and algorithms & CTR & MRR\\
    \midrule
         Use a single recommended item with TRPO & 0.33 (at 30k step) &/\\
         Use an ordered list of recommended items with TRPO &0.37 &0.48\\
         Use a single recommended item with DQN & 0.30 & /\\
         Collaborative Filtering & / & 0.28\\
         Recommendations based on Popularity & / & 0.29\\
         Random Recommendations & 0.04 & 0.1526\\
    \bottomrule
    \end{tabular}
    \caption {comparison of training results of different algorithms.}
    \label{tab:comp}
\end{table}
We can see that both the TRPO and DQN algorithms have a large improvement relative to random recommendations, but the TRPO algorithm with a single recommendation item performs better and can quickly improve performance with less training, reaching a click-through rate of about 0.35. The following is a step-by-step explanation of the performance of each algorithm.
In the following we will illustrate the performance of each algorithm.

\subsection{Trust Region Policy Optimization}
\begin{figure}[htbp]
    \centering
    \includegraphics[width=0.8\textwidth]{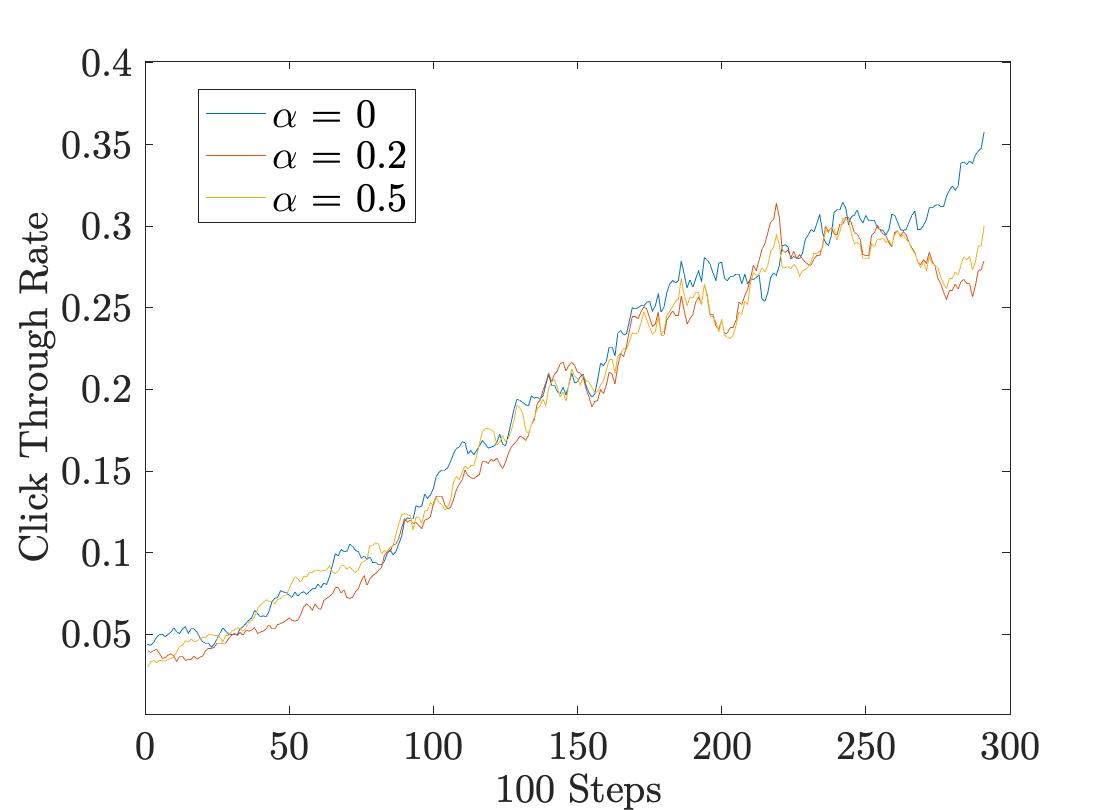}
    \caption {the performance of a single recommended item with TRPO algorithm under different $\alpha$ (using a moving average with interval of length 1500)}
    \label{fig:trpo1}
\end{figure}

The TRPO algorithm can adapt to many different forms of action. We tested the training results of TRPO when the action is a single recommended item, as shown in \ref{fig:trpo1}.
It can be seen that the TRPO algorithm is able to increase the value of the click-through rate faster, reaching a click-through rate of around 0.35 at 30k training steps. There is little difference in the performance of the algorithm when the learning rate $\alpha$ of the user preference vector varies, but $\alpha=0$ have an overall better performance. This result suggests that user preference can be largely determined by the last item the user interacts with.

\begin{figure}[htbp]
    \centering
    \includegraphics[width=0.8\textwidth]{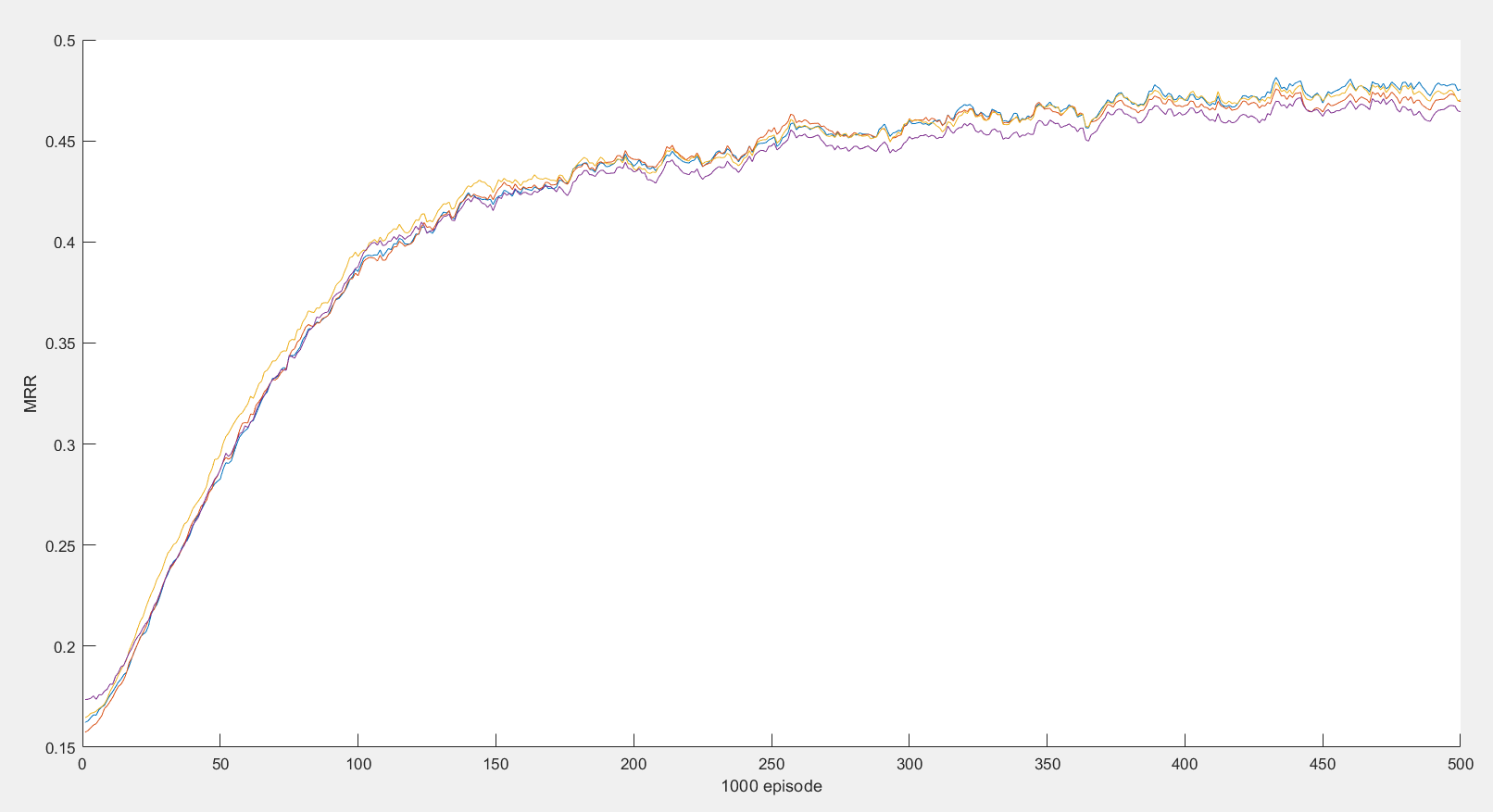}
    \caption {result of training three times with the TRPO algorithm}
    \label{fig:trpo}
\end{figure}

In addition, we used action as a continuous vector of $25\times 1$ dimension to represent the user's ratings of the 25 hotels. By rating, we ranked the 25 hotels and returned the average reciprocal ranking as rewards. The training results are shown in Figure \ref{fig:trpo}.

Figure \ref{fig:trpo} is the result of training 3 times while keeping all the conditions the same. As you can see TRPO is a very stable algorithm that keeps improving the results until the MRR values stabilize.

\begin{figure}[htbp]
    \centering
    \includegraphics[width=0.8\textwidth]{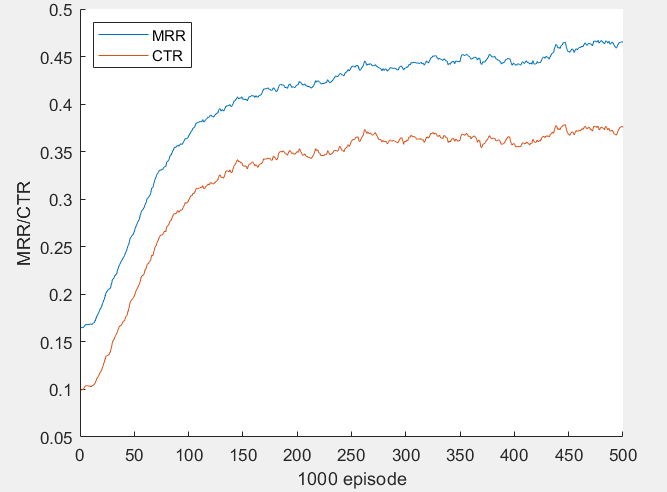}
    \caption {click through rate (CTR) vs. average reverse ranking (MRR) performance}
    \label{fig:ctr}
\end{figure}

The average reciprocal ranking in Figure \ref{fig:ctr} is a very similar training curve calculated using the same method as Figure \ref{fig:trpo}, while the orange line is a click-through rate curve to compare to other methods that may not have an average reciprocal.

In general, the TRPO algorithm have more stable performance, and can achieve a faster increase in click-through rate. However, in our program, the training of the TRPO algorithm is slow, so there is room for further optimization.

\subsection{Deep Q network}
\begin{figure}[htbp]
    \centering
    \includegraphics[width=0.8\textwidth]{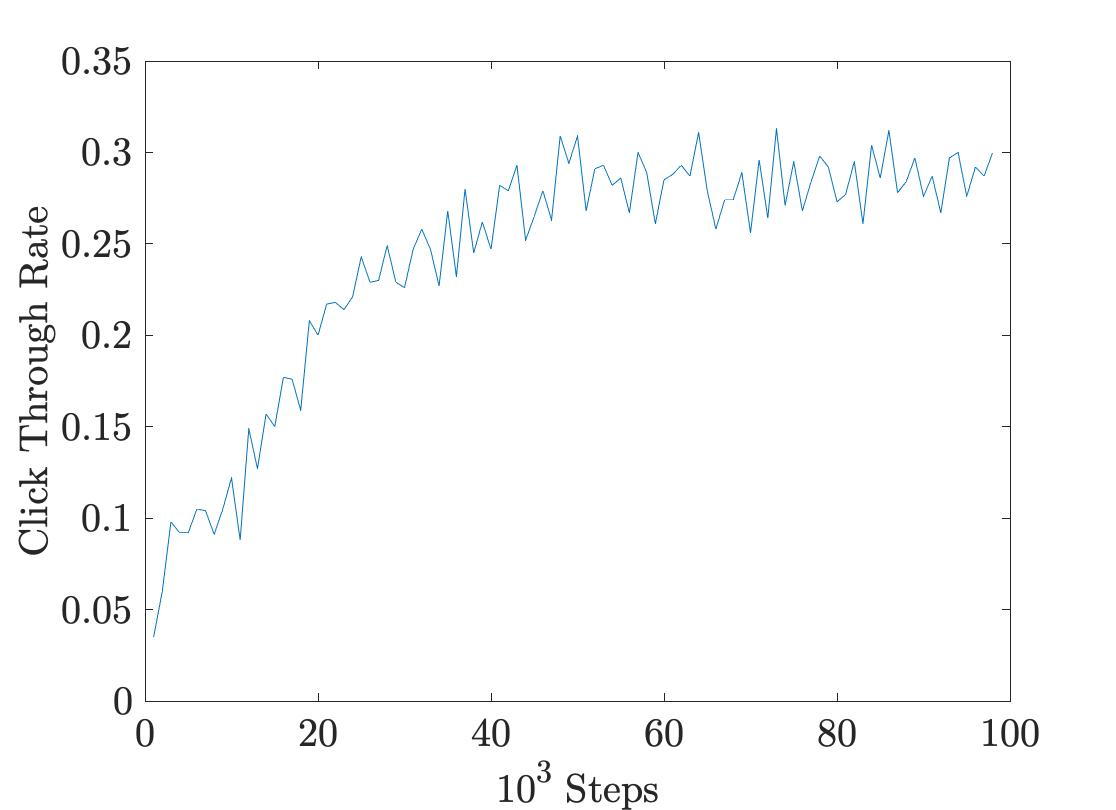}
    \caption{DQN training results.}
    \label{fig:dqn}
\end{figure}

Using the single-item recommendation method and the Deep Q Network (DQN) for training, we were able to obtain the training results shown in \ref{fig:dqn}.
We can see that the DQN can achieve a click-through rate of about 0.29 after 50k training episodes. However, examining the training results shows that in the later stages of training, the DQN often just returned the item that was ranked first among the candidate items as a recommendation, which did not show the advantage of reinforcement learning.

\section{Reinforcement Learning based Recommender System using Biclustering Technique}
\subsection{Introduction}
Though we successfully build an environment for reinforcement learning in the last section, it is not always the case that users may change their preference during browsing items. Actually, if a user had only looked at a few items, he is unlikely to change his preferences. This usually happens when a user has just entered the recommender system. Therefore, in some cases the supervised learning algorithm is still important. Therefore, we want to find an algorithm which combines the advantages of both supervised learning and unsupervised learning.

Nowadays. the most successful supervised algorithm used in recommender system is collaborative filtering (CF). The core idea of CF is homophily of groups. That means users who like the same item are similar in some features. In the same way, items which are prefered by the same user will have some similar features. With this idea, CF can make a personalized prediction about the preferences of users according to the preferences of other users who have interest in the same items. However, there are still some disadvantage of CF. First of all, most CF are user-based or item-based, which consider only one dimension out of two. This may lead to the loss of information. 

In this section, we found a newly proposed reinforcement learning algorithm in recommender system \cite{rlbiclustering}. It makes use of the advantages of biclustering, which could consider features of user list and item list together, and use reinforcement learning to capture state transition of users. However, the grid world they used limits the recommendation path space. This section is structured as follows. In Section 9.2, we will introduce the biclustering reinforcement learning algorithm we want to improved. In Section 9.3 we will point out the disadvantage of this algorithm. In Section 9.5, we will show our proposed algorithm. In Section 9.5, we show our experiment results to prove that our algorithm successfully improves the performance. In Section 9.6 we discuss the conclusion and future works.
\subsection{Original Algorithm Description}
\subsubsection{Introduction to Biclustering}
Given a matrix, Bimax biclustering can be used to capture highly related biclusters \cite{bimax}. For example, in a user-item mtrix $S$, a specified position $S_{u,i}$ stands for whether user $u$ has chosen item $i$. If $u$ choose $i$, the value will be 1, otherwise the value will be 0. Biclustering could find special matrices, which is called biclusters, their x-axis is a subset of user set, and y-axis is subset of user set. In this matrix, all values are equal to one, which means that all users in x-axis have chosen all items in y-axis. Therefore, every sub-matrix stands for a kind of items, which are prefered by a kind of users.

\subsubsection{Construction of State Space}
This algorithm builds a user-item matrix which mentioned in last subsection based on movie-lens dataset. In this matrix, the ratings of user range from 1 to 5.  Therefore, we need to map the rating of users into 0 or 1. We set a threshold to this mapping. If the rating is less than the threshold, than it is mapped to 0, which means the user dislikes the item. If the rating is larger or equal to the threshold, then it is mapped to 1, which indicates that the user may like the item. Then, biclustering is applied to produce $n^2$ small matrices. 

Then, these $n^2$ matrices will be arranged into a grid world with length $n$. The arrangement is based on the distance between user vectors in each matrix. By using a greedy algorithm which minimizing the total distance, the $n^2$ matrices are mapped into a grid world. This grid world constructs a two-dimensional euclidean space. With such a two-dimensional euclidean space, the recommender agent will move in the grid world, and every position will have a coordinate. Then, we will use reinforcement learning to learn policy of the agent.
\subsubsection{Definition of Action Space, Transition Function and Reward}
The coordinate of the agent is its current state, and the coordinate of the agent after action is next state. The state is denoted as $s = (X,Y)$, where $x$ is the x-axis of euclidean coordinate and $y$ is the y-axis of euclidean coordinate.
In a two-dimension euclidean space, there are four actions for the agent to move: up, down, left, right. Therefore, the action space is defined as a discrete four dimensional space.
The gridworld is deterministic, which means that the transition function is relatively fixed. The transition function is determined by current position and action, such as $T((x,y),down)=(x+1,y)$.
Reward function $R(s,a,s')$ is also deterministic and it is determined by the similarity of user sets, as follows:

\begin{align*}
R(s,a,s')&=Jaccar\_Distance(U,U')\\
&=\frac{\left|U\cap U'\right|}{U\cup U'}
\end{align*}
where $s$ and $U$ indicate the current state and user vector in current state, and $s'$ and $U'$ indicate the next state and user vector in next state.

\subsubsection{Interaction with Recommender Agent}
This algorithm uses Q-learing agent to determine action policy. When this Q-learning agent move during the gridworld, each action will bring a reward, which is given according to the reward function is last subsection. Since the gridworld is deterministic with limited number of state, and reward function only relates to two adjacent grids, it is a good choice to use Q-learning agent to determine the policy.

When the recommendation agent enters a new state, this algorithm will recommend all items included in the item list of the corresponding bicluster of this state. For each user, this algorithm will build a list to restore all recommendations to this user, which is called recommendation list. This list has a limitation, which is denoted as N. When this recommendation agent enters a new state, and all items in the bicluster of this state are included in the recommendation list, the interaction will end.

For each user, this algorithm will choose $m$ starting positions according to the history record of the user. When the interaction of one starting position ends, it will start from a new starting position. When interactions with all starting positions end or the recommendation list reaches its limitation, the interaction with this user will end.
\subsection{Existing problem}
The original algorithm better extracts the characteristics of both users and items, but we also found some disadvantages. One disadvantage is that the gridworld constructed by this algorithm is not stable. This gridworld is constructed by minimizing the total distance between states, and the greedy algorithm it uses can only ensure that it converges to a local minimum, but not a global minimum. In this case, the recommended accuracy of this algorithm will be affected by the convergence result. 

Another disadvantage is that the possible recommended path of this algorithm is limited by the gridworld structure. Suppose that two items favored by a user, but these two items are distributed in two corners of the board. This means that if you want both items to be recommended, the agent needs to travel through the entire board. However, the recommendation may have been stopped because no new items were found, so items that need to be recommended may not necessarily be recommended.
\subsection{Proposed Solution}
\subsubsection{Increasing stability of the gridworld}
To increase the stability of this gridworld, we plan to use simulated annealing algorithm to replace greedy algorithm. First of all, we will arrange all biclusters into the gridworld. Then, we set sum of distance between all adjacent grids as a heuristic function $h(s)$ as shown below.
\begin{algorithm}[!htbp]
		\caption{Computation of h(s)}
		\SetKwInOut{Input}{Input}
    	\SetKwInOut{Output}{Output}
		\Input{set of all vertices $V$, current gridworld arrangement state $S$.}
	    \Output{total distance between adjacent vertices $h(s)$}
		$h(s) \leftarrow 0$;\\
		\For {$v$ in $V$}
		{
			\For {$v'$ in  neighbor(v)}
			    {
				$h(s)\leftarrow h(s) + distance(v,v') $;
			    }
		}
		 \Return{$h(s)$};	
		
\end{algorithm}
Then, we use simulated annealing to minimize this algorithm. According to the result of simulated annealing, we will map all biclusters into the gridworld. Since this is a NP-hard problem, we may not be able to find a global extreme even we use simulated annealing. However, if the execution time is long enough, this algorithm can provide a better solution than greedy algorithm.
\subsubsection{Increasing State Space and Action Space}
Our main task focuses on increasing the behavior space and enriching the recommended paths. To achieve this, we decided to use multiple boards to build the state space. We use K as notation of the number of boards that we use. First, we obtain K boards via running K times simulated annealing algorithm. Since it does not converge to the optimal solution and the starting states are different, the two executions of the algorithm will yield different K boards. However, we are using the same $n^2$ small submatrices, and thus these K boards will only differ in the order of nodes and contain the same information. In other words, for any point $v$ on each board, you can find points on other boards that contain the same user and item lists with different positions. For any point $v_1$ on the first board, there will be another k-1 identical points distributed on the other k boards, marked with bits $v_2$, $v_3$, $v_4$, ..., $v_k$. Mark their position information as ($x_1$, $y_1$), ($x_2$, $y_2$), ... , ($x_k$, $y_k$). Then, we construct the new behavior space as s=\{($x_1$,$y_1$), ($x_2$,$y_2$),..., ($x_k$,$y_k$)\}. This does not change the total number of state spaces, so the Q-Learning algorithm still applies to determining strategies.

	The improvements brought about by the addition of the new board are mainly in the behavior space. On each board, each node provides 4 possible behavioral decisions. Then, when we combine these k nodes to form a state, the number of possible behavioral decisions for this state increases to 4k. This method can arbitrarily expands the behavior space, enhancing the connectivity of the recommendation map, and theoretically reducing the impact of simulated annealing convergence.
	
	For example, in the original algorithm, for the same user, the two items that need to be recommended are in submatrices on the two side of the board (far apart), which requires a long path to connect them. In our algorithm, the two items have a distance between them in each board, and we only need to connect the shortest path, so the probability that they will be recommended together will be greatly increased.

\subsection{Experiment}

\subsubsection{Dataset}
We use the interfaces and classes provided by the OpenAI Gym to build our environment and then use the self-built Q-Learning agent to make behavioral decisions. In addition, to remain consistent with the original paper, we  use the Movielens dataset to measure our algorithm. The data set contains 100,000 scores, and having a 3-point threshold, i.e., 0 for a score of three or less, and 1 for a score of three or more, as previously mentioned. 80 percent of the dataset will be used as a training set, and the rest will be used as a test set. In order to evaluate the performance of our algorithm under cold start condition, only one-tenth of user's historical data is left as observable when testing.
\subsubsection{Initial setup}
To be consistent with the original paper, we use the Bimax biclustering algorithm to process the raw data. We randomly select $n^2$ small matrices to simulate the annealing algorithm. In the experiment, the length n of the board is set to be 20. We use the Q-learning algorithm to learn strategy.
During the experiment, to verify the performance of our algorithm in recommending different number of items, we vary the length of the recommended list for testing recall rate.
\subsubsection{Experiment Result}
We evaluate our algorithm by the recommendation recall, which measures the ability of the recommendation system to find out all the favorite items of users. They are defined by the following formula.

$$Recall = \frac{\left|I_{N}\cup I_{D}\right|}{N}$$

\begin{figure}[H]
    \centering
    \includegraphics[width=0.8\textwidth]{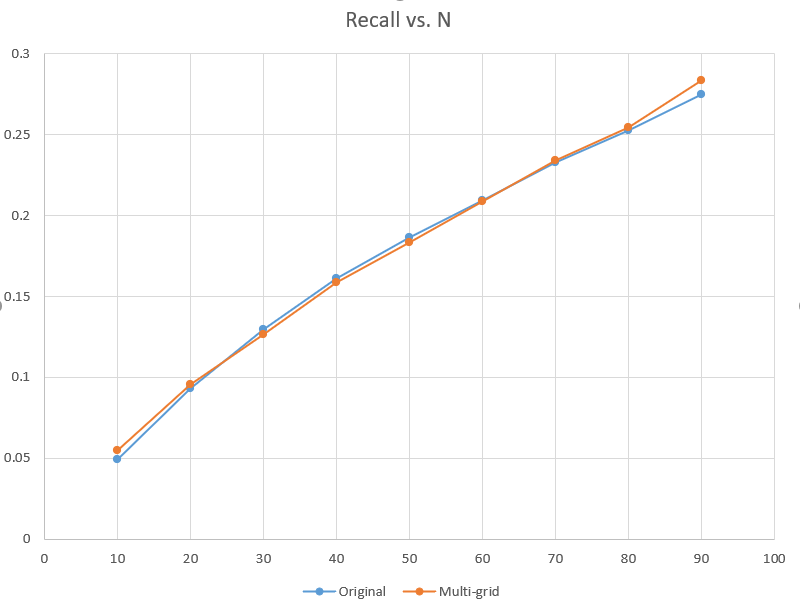}
    \caption{Recall vs. N}
    \label{datum}
\end{figure}

Figure 10 shows the final experimental results. We compare the user-based collaborative filtering algorithm, the original biclustering reinforcement learning algorithm and our proposed algorithm. However, this experimental result is not satisfying. As we can see from the figure, the revised algorithm is nearly identical to the original algorithm in terms of recall rate. In the comparison of the recall rate, we have a slight improvement over the original algorithm when N is less than or equal to 20 or greater than or equal to 70, but our algorithm is beaten between 30 and 60. After analysis, we guess that this is due to our inappropriate reward settings for the Q-Learning agents. We set the reward as the similarity of two neighboring points according to the original algorithm, which is fixed. However, for a given user, this similarity does not represent the actual value of the two points. Specifically, suppose we have a user whose record is $U$, our recommendation agent is represented by being at a point $a$ and having two points $b$ and $c$ adjacent to $a$. Even if we know that $Similarity(a,b)>Similarity(a,c)$, we cannot determine $Similarity(U,b)>Similarity(U,c)$. In addition, although positively correlated, similarity is different from user click-through rate. As we do not directly reward click-through or recall rate, an increase in recall is not guaranteed. We guess that both our improved and original algorithms have reached their limits of the reward mechanism, thus producing nearly identical results.

However, we believe that there is still room for improvement in our algorithm. Theoretically, the set of recommended paths of our algorithm contains those of the original algorithm, so the recommended paths generated by our algorithm are at least not weaker than those of the original algorithm. With a more scientific reward mechanism, we may find better recommended paths. In addition, our algorithm has good generalization. As long as there is some way to package the items as nodes, we can apply this algorithm to provide a recommendation map with high connectivity and good maintainability.

\section{Conclusion}
In this IPP project, we explored ways to combine the recommendation system with reinforcement learning.

First, we attempted to create a reinforcement learning environment for linking historical datasets to reinforcement learning. In this process, we used data normalization, exponential moving average, and other methods, and tried a number of different combinations of action, reward and reinforcement learning algorithms. In the end, we found that the TRPO algorithm with a single recommendation performs better and more stable, and can improve the click-through rate to about 0.35 in just 30k training steps (user interactions).

Secondly, we try to improve an algorithm based on bi-clustering and reinforcement learning to expand the action space and increase the connectivity of the recommendation map. However, due to an unreasonable reward mechanism, the results are not satisfactory and the results of our algorithm are similar to the original algorithm.

There is still much room for improvement in our work.
\begin{enumerate}
    \item The training of our reinforcement learning environment using a single-item recommendation with TRPO algorithm is very time consuming and has a lot of room for acceleration.
    \item In setting up the reinforcement learning environment, we were unable to address the existing problems of Virtual Taobao. Since our user information and user actions are obtained directly from the dataset, the recommendations made by the reinforcement learning algorithm only affect the reward of this episode, but not the user's next action (performing the next search or leaving directly). Solving this problem can make the Markov chain more complete and rational.
    \item Theoretically, our improved bi-clustering with reinforcement learning algorithm accommodates the recommendation paths of the original algorithm, and it will be possible to find more optimal recommendation paths if a more reasonable reward mechanism is found.
    \item Our improved bi-clustering with reinforcement learning algorithm is generalizable and can provide a recommendation map for clustered items that has both high connectivity and simplicity.
\end{enumerate}

\end{document}